\documentclass[sigconf]{acmart}
\usepackage{hyperref}
\usepackage{graphicx}
\graphicspath{ {./images/} }
\usepackage{multirow}
\usepackage{amsmath}
\usepackage{rotating}
\usepackage{xcolor}
\usepackage{subcaption}
\usepackage{adjustbox}
\usepackage[capitalize]{cleveref}
\usepackage{balance}
\crefname{section}{Sec.}{Secs.}
\Crefname{section}{Section}{Sections}
\crefname{table}{Tab.}{Tabs.}
\Crefname{table}{Table}{Tables}
\usepackage[outdir=./images/]{epstopdf}

\AtBeginDocument{%
 }

\copyrightyear{2024}
\acmYear{2024}
\setcopyright{rightsretained}
\acmConference[HRI '24 Companion]{Companion of the 2024 ACM/IEEE International Conference on Human-Robot Interaction}{March 11--14, 2024}{Boulder, CO, USA}
\acmBooktitle{Companion of the 2024 ACM/IEEE International Conference on Human-Robot Interaction (HRI '24 Companion), March 11--14, 2024, Boulder, CO, USA}
\acmDOI{10.1145/3610978.3640648}
\acmISBN{979-8-4007-0323-2/24/03}

\makeatletter
\gdef\@copyrightpermission{
  \begin{minipage}{0.3\columnwidth}
   \href{https://creativecommons.org/licenses/by/4.0/}{\includegraphics[width=0.90\textwidth]{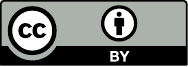}}
  \end{minipage}\hfill
  \begin{minipage}{0.7\columnwidth}
   \href{https://creativecommons.org/licenses/by/4.0/}{This work is licensed under a Creative Commons Attribution International 4.0 License.}
  \end{minipage}
  \vspace{5pt}
}
\makeatother

\begin{document}

\title{Improving Visual Perception of a Social Robot for Controlled and In-the-wild Human-robot Interaction}


\author{Wangjie Zhong}
\affiliation{%
  \institution{Faculty of IT, Monash University}
  \city{Melbourne}
  \country{Australia}}
\email{wzho0026@student.monash.edu}

\author{Leimin Tian}
\affiliation{%
  \institution{Faculty of Engineering, Monash University}
  \city{Melbourne}
  \country{Australia}}
\email{leimin.tian@monash.edu}

\author{Duy Tho Le}
\affiliation{%
  \institution{Faculty of IT, Monash University}
  \city{Melbourne}
  \country{Australia}}
\email{tho.le1@monash.edu}

\author{Hamid Rezatofighi}
\affiliation{%
  \institution{Faculty of IT, Monash University}
  \city{Melbourne}
  \country{Australia}}
\email{hamid.rezatofighi@monash.edu}


\renewcommand{\shortauthors}{Wangjie Zhong, Leimin Tian, Duy Tho Le, \& Hamid Rezatofighi}

\begin{abstract}
Social robots often rely on visual perception to understand their users and the environment. Recent advancements in data-driven approaches for computer vision have demonstrated great potentials for applying deep-learning models to enhance a social robot's visual perception. However, the high computational demands of deep-learning methods, as opposed to the more resource-efficient shallow-learning models, bring up important questions regarding their effects on real-world interaction and user experience. It is unclear how will the objective interaction performance and subjective user experience be influenced when a social robot adopts a deep-learning based visual perception model. We employed state-of-the-art human perception and tracking models to improve the visual perception function of the Pepper robot and conducted a controlled lab study and an in-the-wild human-robot interaction study to evaluate this novel perception function for following a specific user with other people present in the scene.
\end{abstract}



\maketitle

\section{Introduction}

\begin{figure*}[tb]
    \centering
    \begin{subfigure}[t]{0.5\textwidth}
        \centering
        \includegraphics[height=5cm]{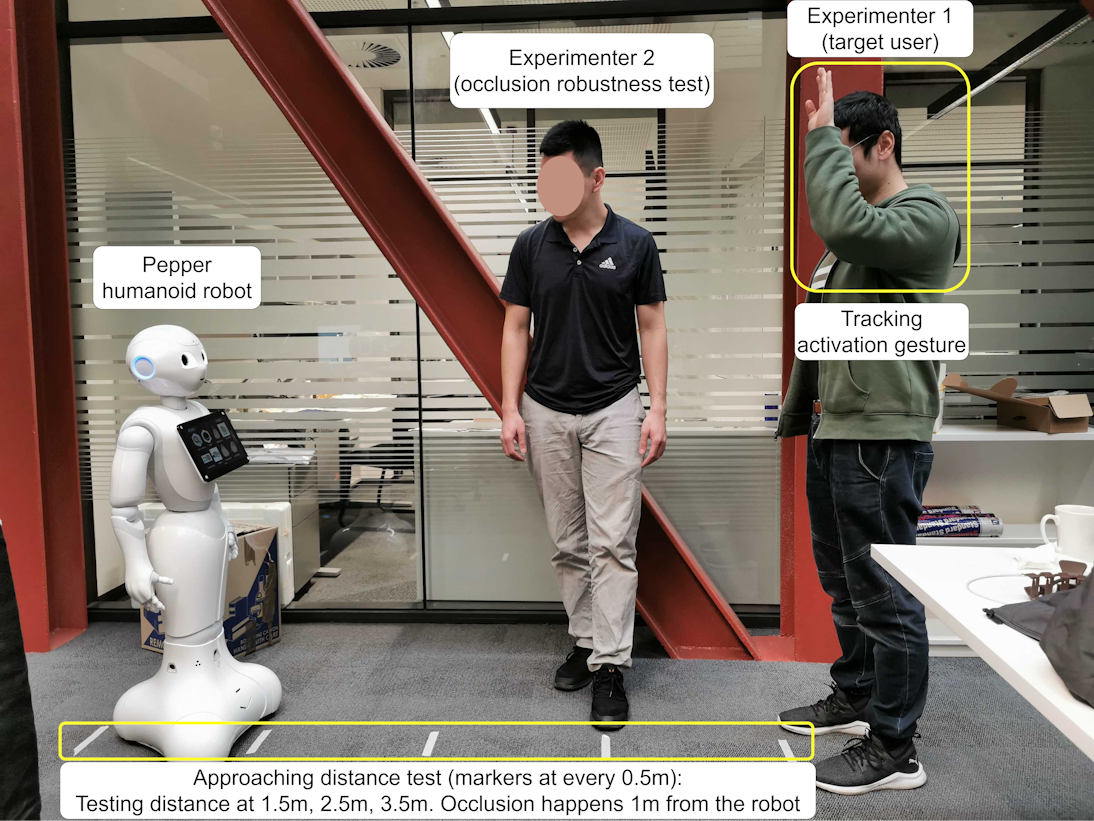}
        \Description{A Pepper humanoid robot is to the left of the photo, a researcher assuming the role of a user is to the right, in the middle another researcher assumes the role of a passer-by. The user raises their hand to activate Pepper's tracking mode. After the robot starts moving towards the user, the passer-by walks in between the robot and the user to create obstruction. On the floor there are masking tapes, which marks the three distances between the robot and the user when the tracking activates, namely at 1.5m, 2.5m, and 3.5m.}
        \caption{Lab-based trials assessing tracking distance and occlusion}
        \label{fig:lab}
    \end{subfigure}%
    ~ 
    \begin{subfigure}[t]{0.5\textwidth}
        \centering
        \includegraphics[height=5cm]{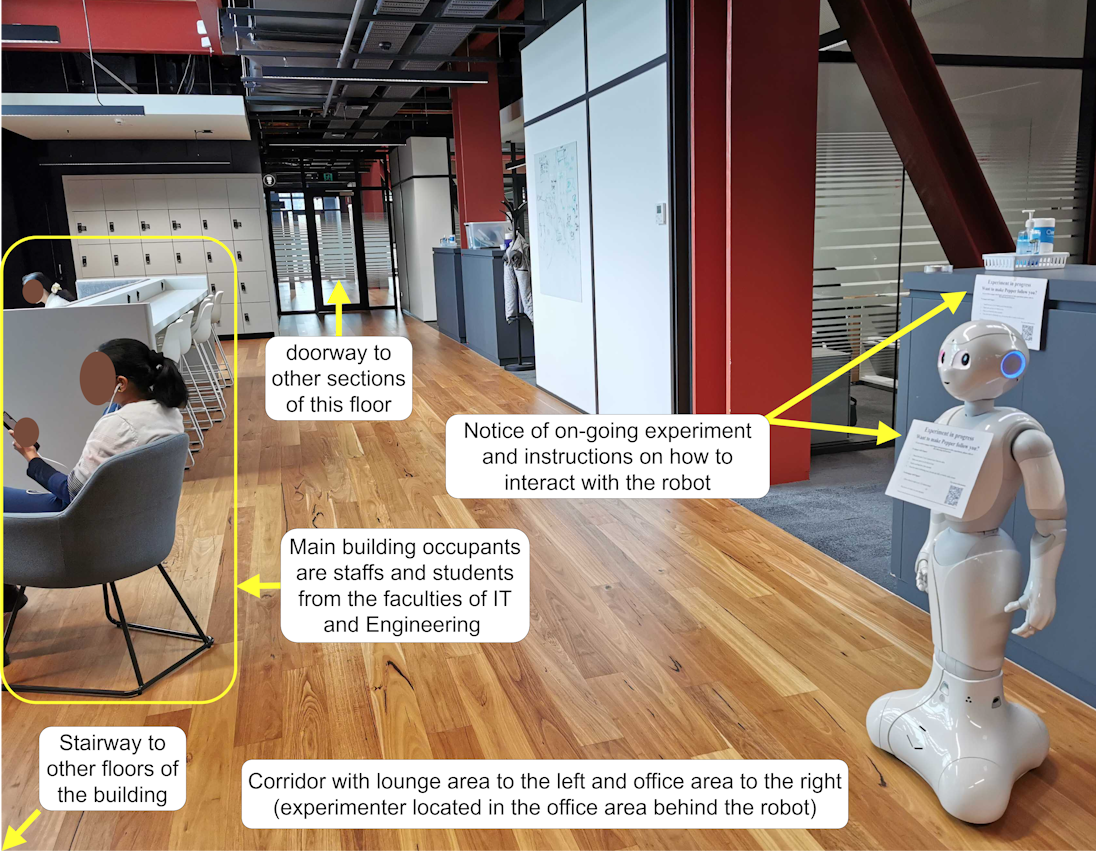}
        \Description{The photo shows the hallway of a university building, with a Pepper humanoid robot to the right of the photo and a person sitting on a chair to the left of the photo in the open lounge area. Behind the robot there is a paper notice hanging on the wall instructing on the ongoing experiment.}
        \caption{In-the-wild trials focusing on objective performance and subjective user experiences}
        \label{fig:wild}
    \end{subfigure}
    \vspace{-1em}
    \caption{Evaluation of the proposed visual perception system}
    \label{fig:exp_layout}
    \vspace{-1em}
\end{figure*}

Robots have been integrated into our society, automating tasks and enhancing human capabilities~\cite{kabacinska2021socially,liang2021human}. To effectively engage with humans, social robots require robust visual perception~\cite{anzalone2015evaluating}, which remains challenging especially in real-world scenarios and public spaces with diverse user and environmental contexts~\cite{bauer2019refining,bourguet2020impact}. 
This study aims to improve human perception capabilities of social robots, demonstrated on the Pepper humanoid robot as a representative commonly adopted in research and application~\cite{pandey2018mass, tuomi2021spicing, tanaka2015pepper}. In particular, we focus on integrating state-of-the-art (SOTA) deep learning (DL) models for 2D multiple human pose estimation and tracking. DL-based computer vision holds the potential to enable robots to have a robust understanding of non-verbal human expressions, including body language, while identifying individuals of interest in a crowded public space~\cite{mirani2022object, wang2022yolov7, maji2022yolo}. 

By equipping social robots with improved human perception capabilities, the objective task outcomes and subjective user experiences of human-robot interaction (HRI) are expected be improved~\cite{jimenez2023deep,she2018improving}. In this work, to investigate the influence of a robot's visual perception capability enhanced by employing SOTA DL models in HRI, we examine an example application scenario in which a social robot tracks a specific user in a public space where other people are also present in the scene, both in controlled lab-based trials and as an in-the-wild study, as shown in \cref{fig:exp_layout}.
Our main contributions are:
\begin{itemize}
    \item We implemented an open-sourced visual perception framework for social robots using SOTA human tracking models and demonstrated its incorporation on the Pepper robot;
    \item We conducted lab-based and in-the-wild evaluation demonstrating improved performance by the proposed framework.
\end{itemize}

\section{Background}
Data-driven approaches have revolutionised computer vision~\cite{li2021low, galab2021adaptive, stiebel2020brightness}. When integrated with robotics, DL models yielded significant contributions to HRI, enabling robots to recognize and interpret human cues such as body poses and hand gestures~\cite{robinson2023robotic}. 
Human pose estimation involves detecting the body poses of individuals by identifying keypoints that represent important joints~\cite{chen20222d}. There are two common approaches to pose estimation: top-down and bottom-up.
The top-down approach detects humans by finding bounding boxes of individuals and then estimating the keypoints within each bounding box~\cite{nguyen2022survey,kulkarni2023poseanalyser}. It has achieved SOTA performance, however the computation scales with the number of detected humans and it struggles with occlusion~\cite{zhang2019pose2seg}. The bottom-up approach estimates all keypoints and achieves real-time inference, however it requires complex post-processing to group keypoints for the same person~\cite{dang2019deep}.
Hybrid approaches combine top-down and bottom-up methods~\cite{li2019multi,li2018bottom,zhao2023dpit}, offering faster and more reliable predictions of occluded keypoints. One notable SOTA model is YOLO-Pose~\cite{maji2022yolo}, which predicts the location of obscured joints by estimating a fixed number of keypoints in each bounding box. 

Tracking individuals across frames and scenes by identifying and assigning unique identifiers to detected individuals is crucial in complex HRI tasks~\cite{luo2021multiple, han2020complementary}.
Online tracking methods adopting tracking-by-detection demonstrated leading performance on various benchmark tasks. Specifically, ByteTrack is a generic association method compatible with existing trackers and suitable for situations with limited computational resources~\cite{zhang2022bytetrack}; 
BoT-SORT addresses unpredictable camera motion and is desirable for mobile robots~\cite{aharon2022bot}; OC-SORT significantly improves tracking quality through observation-centric recovery, enabling effective tracking in crowded environments with occlusion~\cite{cao2022observation}. 
We tested deploying these SOTA tracking methods on a social robot to investigate their benefits for a robot's visual-based human perception in HRI.

We chose Pepper as a representative social robot adopted in research and application~\cite{pandey2018mass}. While it offers basic human perception, the models offered by the onboard NAOqi library remain proprietary. 
Existing approaches to enhance Pepper's visual perception connected the robot's input and output functions with other visual perception libraries, such as the Robot Operating System. However, existing work has not integrated SOTA online human tracking and pose estimation, limiting its visual perception performance in complex scenarios~\cite{caniot2020adapted,augello2020kinect}.


\section{Methodology}

\subsection{Visual Perception Framework}

\begin{figure}[tb]
    \centering
    \includegraphics[width=\linewidth]{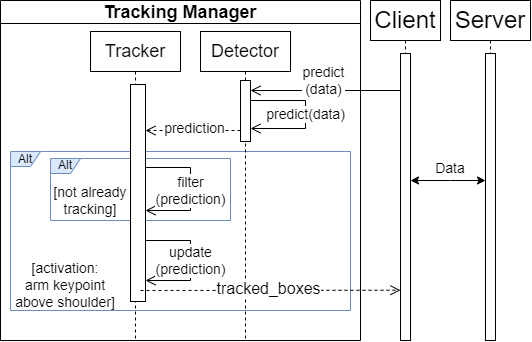} 
    \vspace{-1em}
    \Description{The visual perception component contains a tracking manager that includes the tracker and detector modules, as well as the client and the server which transmits data in between. The client sends data to the detector, which sends prediction to the tracker. The tracker updates the prediction and sends the tracked boxes back to the client.}
    \vspace{-1em}
    \caption{The visual perception components for user tracking.}
    \label{fig:prediction}
    \vspace{-1em}
\end{figure}

\begin{figure}[tb]
    \centering
    \includegraphics[width=\linewidth]{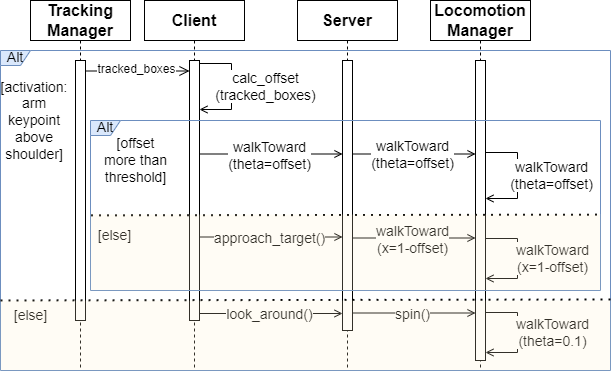} 
    \vspace{-1em}
    \Description{The behaviour flow of user following contains a tracking manager, a client, a server, and a locomotion manager. The tracking manager sends tracked boxes to the client, which sends the approach target or look around commands to the server. The server communicates with the locomotion manager to execute the walk-towards-user action to achieve user following.}
    \vspace{-1em}
    \caption{Following the user after receiving tracking data.}
    \label{fig:follow}
    \vspace{-1em}
\end{figure}

To integrate SOTA computer vision models with Pepper, we developed a Python framework\footnote{All materials: \url{https://github.com/Fresh-Broccoli/pepper_DL/tree/pepper_skeleton}}. This framework establishes a server-client architecture to overcome compatibility issues between the robot's function library and SOTA computer vision libraries. The server streams live footage from Pepper's front camera to the client for processing, then the client generates suitable robot actions in response, which are sent to the server for Pepper to execute. This architecture facilitates easy adaption to different DL models on the client side and different robot platforms on the server side 

\subsection{Following a Specific User}

To investigate the proposed framework for enhancing human perception in HRI, we designed an interaction scenario in which the robot tracks and follows a specific user who raises their hand while there are other humans present in the scene who may walk in between the robot and the target user during tracking. \cref{fig:follow} shows an overview of the robot behaviour flow. This is designed to replicate a common HRI scenario happening in public spaces, such as shops or restaurants. We utilised YOLO-Pose~\cite{maji2022yolo} and OC-SORT~\cite{cao2022observation} to perform keypoint estimation and human tracking respectively, facilitating target identification and tracking. Additionally, we explored two alternative SOTA tracking algorithms, namely ByteTrack~\cite{zhang2022bytetrack} and BoTSORT~\cite{aharon2022bot}.

\subsubsection{Camera Configuration}

We stream inputs from Pepper's top camera (resolution 160x120px, $\approx$10FPS) via wireless connection to a controller computer running the visual perception framework.

\subsubsection{Identifying a Specific User}
Tracking activation is determined by calculating the difference between the wrist keypoint and the shoulder keypoint estimated by the pose detection module in our proposed framework. If the wrist is above the shoulder, the robot assumes that the individual is raising their hand deliberately. To reduce false triggers, tracking is initiated only if the hand remains raised for multiple consecutive frames. If multiple people raise their hand simultaneously, the robot selects the individual with the highest predicted confidence. Once a specific user is identified, all other detections are filtered out. The tracker processes this information and assigns an ID to the target user to ensure an accurate identification amongst detected individuals throughout the interaction.

\subsubsection{Tracking Activation and Approaching A User}

Once tracking is activated, the robot notifies the user with a spoken phrase ``target detected'' and begins centering them in its camera view by rotating its body. The angular velocity is calculated as $\frac{x_{\text{Frame Centre}} - x_{\text{Box Centre}}}{\text{Frame Width}} \cdot n$ (positive value represents anti-clockwise rotation, i.e., if the target is to the robot's right it rotates to its right):
\begin{itemize}
    \item $x_{\text{Frame Centre}}$ is the width-wide centre of the frame
    \item $x_{\text{Box Centre}}$ is the width-wide centre of the bounding box
    \item \textit{Frame Width} is the width of images captured by Pepper
    \item $n$ is an adjustment constant set to 0.9, i.e., slowing the default angular velocity by 10\%
\end{itemize}

Once the user's x position is within a predefined threshold, Pepper approaches them at a linear velocity inversely proportional to the ratio of the bounding box area to the entire frame area. This allows Pepper to move faster when the user is far away and slower when they are close. The stopping condition is reached when the bounding box exceeds a certain ratio within the robot's view, i.e., the robot reaches a pre-defined distance to the user. After stopping, the robot asks ``Hello, I'm Pepper, do you require my assistance?'' to conclude tracking.

\subsubsection{Recovery from Occlusion}
We tested the proposed framework's resilience against occlusion. The tracker uses a custom patience threshold to resume tracking if the target is being temporarily obscured. To resume tracking of a mobile target, our framework stores a copy of the target's last bounding box. Based on this information, it determines the most likely direction in which the target has moved during occlusion. For example, if the target's bounding box was on the left side of the screen, the robot will turn left, anticipating the target's user's potential movement and new location.

\subsection{Default Tracking with NAOqi}

We compared the proposed framework with Pepper's default NAOqi 2.5 SDK, namely the \href{http://doc.aldebaran.com/2-5/naoqi/peopleperception/alpeopleperception.html}{ALPeoplePerception} module. This function divides the area in front of Pepper into three engagement zones based on 3 distances: Zone 1 at 0-1.5m, Zone 2 at 1.5-2.5m, and Zone 3 at beyond 2.5m, with the maximum effective range of Pepper's 3D camera, ASUS Xtion, at 3.5m~\cite{trevor2012planar}.


\section{Experimental Protocols}
The experimental protocols have been reviewed and approved by the university's human research ethics committee (ID 39135). We conducted lab-based trials focusing on evaluating the performance of the proposed framework using DL-based human perception models compared to the default human perception function of Pepper. Further, we conducted an in-the-wild study in which passers-by are free to interact with the robot and evaluate the performance of the proposed framework using observations and questionnaires.

\subsection{Lab-Based Trials}
We evaluated the range and robustness of the proposed framework against occlusions in a controlled indoor office environment with consistent lighting conditions (\cref{fig:lab}). The trials were conducted using a controller computer with an Intel i9-10900k processor, NVIDIA RTX 3090 GPU, and 64GB RAM. We used default parameters in all SOTA human detection and tracking models. 

We tested three distances between the user and the robot at tracking activation, namely at 1.5m, 2.5m, and 3.5m, corresponding to Pepper's engagement Zones. At each of these activation distances, we compared 4 human perception models for the tracking behaviour, namely ByteTrack, BoT-SORT, OC-SORT, and the default non-DL model offered by NAOqi. For each distance and model, we ran 10 trials (120 trials in total). In these trials, one experimenter assumed the role of the user. Another experimenter assumed the role of a passer-by who walked in-between the robot and the user during tracking at 0.5m in front of the user to cause occlusion.

The robot was allowed 30 seconds to activate tracking from the moment the user raised their hand. A trial is considered a success if Pepper reached the user, and considered a failure if Pepper failed to activate tracking, lost track of the user, or followed the passer-by instead. In addition to trial success and failures, we recorded trial footage, measured the time taken to complete each trial, and the number of occluded frames. 

\subsection{In-the-Wild Study}
To further understand the outcomes of the proposed perception framework, we deployed Pepper in an open lounge area in a university building near a stairway (\cref{fig:wild}). The occupants are mainly staff and postgraduate students from the faculties of IT and Engineering. Using our DL-based behaviour, we ran four 2-hour sessions during weekdays at noon (8h deployment time in total). The robot was left alone and interaction with it was entirely voluntary. Signs were posted on and around the robot's surroundings detailing instructions for triggering our DL behaviour, as well as a QR code for filling in an evaluative questionnaire (perceived intelligence, perceived safety~\cite{bartneck2009measurement}).

An experimenter supervised the sessions from a nearby office and collected observation notes (\cref{tbl:obs_data_description}). We counted the number of passer-by or foot-traffic and each passer-by constitutes one row in the observation notes. We used OC-SORT for human tracking in the deployment, as it yielded good performance for occlusion robustness in the lab-based trials. Due to network connection restrictions, we used a different controller computer (Intel i7-12800H processor, NVIDIA RTX 3080 GPU, 64GB RAM).

\begin{table}[h!]
    \centering
    \begin{adjustbox}{width=\columnwidth,center}
    \begin{tabular}{l|p{1.2cm}|p{5cm}}
    \toprule
        \textbf{Label} & \textbf{Values} & \textbf{Description} \\
    \midrule
        Participant & 1 or 0 & Attempted to interact with the robot \\
        Survey & 1 or 0 & Filled in questionnaire \\
        Success & 1 or 0 & Tracking successful \\
        Invested & 1 or 0 & Interacted with the robot multiple times \\
        Photo & 1 or 0 & Took photo or video \\
        Emotion & Positive, Neutral, Negative & Observable emotions exhibited by the individual in the presence of the robot \\
    \bottomrule
    \end{tabular}
    \end{adjustbox}
    \caption{Description of in-the-wild study's observation notes}
    \label{tbl:obs_data_description}
    \vspace{-2em}
\end{table}

\section{Results}
\subsection{Lab-Based Trials}


As shown in \cref{tbl:count_table}, OC-SORT achieved the highest success rate of 76.7\% averaged over all trials. Further, in all conditions except for at 1.5m distance, the default NAOqi model was outperformed by the proposed DL framework. We observed that ID switch, where the target user's ID was mistakenly assigned to another individual during occlusion was the primary reason for all failed trials, except for the default model at 3.5m, which failed due to activation timeout.
Aggregating the DL model's results, the proposed framework achieved a success rate of 68.9\% compared to the default model with a success rate of 33.3\% ($p = 0.001$ with Fisher's Exact Test). 
We did not find any significant differences between the three DL trackers in terms of success rates or trial duration. BoT-SORT had significantly lower frame rate than ByteTrack and OC-SORT. In terms of the number of occluded frames, i.e., the number of frames where the tracker temporarily lost the target and actively tried to re-establish tracking, OC-SORT has significantly higher mean number of occluded frames than BoT-SORT in successful trials, suggesting that OC-SORT was more robust against occlusion.

\subsection{In-the-Wild Study}
During in-the-wild deployment, we recorded a total foot-traffic of 425 people, 13 interacted with the robot (3.1\% attention rate), 8 filled in the questionnaire afterwards. 84.6\% of the interaction attempts were successful. We recorded 53.9\% of people showing visible positive emotions during their interaction with the robot, 84.6\% of them took photos or tried interacting with the robot multiple times.

\section{Discussion} 
Our lab-based and in-the-wild evaluation demonstrated that the proposed framework adopting SOTA pose estimation and tracking models yields reliable performance in the presence of occlusion and the complexity of public spaces. 

For the lab-based trials, our system demonstrated a significantly higher success rate compared to the default system at various activation distances. Further, we tested three trackers incorporated in the proposed framework. The results showed that ByteTrack and OC-SORT exhibited higher frame rates and better occlusion tolerance than BoT-SORT, making them suitable for a social robot tasked to interact with specific users in public spaces. 
Surprisingly, the default non-DL model was able to handle occlusion to some extent in the lab-based trials, especially at shorter distances. One explanation may be that at 1.5m the robot may not be able to see the occluding person's face and thus registering them as a potential user to follow. We also observed that the default model took longer to activate tracking as the user stood further away from the robot. 

The in-the-wild study further validated the performance of the proposed human perception model. However, it is worth noting that the low rate of passer-bys interacting with the robot suggests that more expressive robot behaviours are needed for engaging users in field deployment. Improving the objective performance of a social robot's visual perception is only the first step towards a better HRI. Further investigations are required to understand additional factors that may influence the objective and subjective outcomes of a social robot providing services in public spaces.

\section{Conclusion} 
\label{sec:conclusion}

We proposed a framework to incorporate advanced visual perception models with existing social robots, exemplified with the Pepper humanoid robot. This framework is open-sourced and can be adapted to other visual perception models and robot behaviours. Through lab-based evaluation, we demonstrated that the proposed framework enhanced the robot's ability to detect and track a specific user, achieving an average success rate of 77\% when tested against occlusion and different detection distances, compared to the existing system achieving an average success rate of 33\%. To demonstrate the compatibility of the proposed framework with different visual perception models, we tested three SOTA tracking algorithms, namely ByteTrack, BoTSort, and OC-SORT. The results suggest that ByteTrack or OC-SORT may be suitable for social robots aimed at serving users in public spaces due to their robustness and higher frame rate. 
\begin{table}[tb]
    \centering
    \vspace{-1em}
    \begin{adjustbox}{width=\columnwidth,center}
    \begin{tabular}{c c c c c}
    \toprule
         \multirow{2}{*}{\textbf{Distance}} & \multicolumn{4}{c}{\textbf{Success rate $\uparrow$ (\%, N=10 in a condition)}} \\
         \cmidrule{2-5}
         & ByteTrack & BoTSORT & OC-SORT & Default \\
    \midrule
        1.5m & 100 & 70 & 100 & 90 \\
        2.5m & 20 & 60 & 60 & 10 \\
        3.5m & 70 & 70 & 70 & 0 \\
        \textbf{Average} & 63.3 & 66.7 & \textbf{76.7} & 33.3 \\
    \bottomrule
    \end{tabular}
    \end{adjustbox}
    \caption{Success rate of trackers in the lab-based trials}
    \label{tbl:count_table}
    \vspace{-2em}
\end{table}
Our in-the-wild study further demonstrated the capabilities of using advanced visual perception models in HRI. However, our observations also indicated that user engagement and subjective experience can be influenced by factors beyond the robot's visual perception capabilities. Thus, HRI researchers are encouraged to conduct further investigations to translate the benefits of an improved robot function evaluated from a robot-centred perspective to better service outcomes evaluated from a user-centred perspective.


\newpage
\bibliographystyle{ACM-Reference-Format}
\balance
\bibliography{references}





\end{document}